\pdfoutput=1
\documentclass[11pt]{article}

\usepackage[]{EMNLP2023}

\usepackage{times}
\usepackage{latexsym}

\usepackage[T1]{fontenc}
\usepackage[utf8]{inputenc}

\usepackage{microtype}

\usepackage{inconsolata}

\usepackage{amsmath}
\usepackage{amssymb}
\usepackage[capitalize,noabbrev]{cleveref}

\usepackage{graphicx}
\usepackage{multirow}
\usepackage{booktabs} 
\usepackage{subcaption}
\usepackage{siunitx}
\usepackage{xurl}  % for auto-breaking url lines
\usepackage{enumitem}  % for setting space between list items
\usepackage{makecell}  % for making multirow cells
\usepackage{arydshln}
\usepackage{hyperref}
\usepackage{cleveref}

\crefformat{footnote}{#2\footnotemark[#1]#3}

\DeclareUnicodeCharacter{266B}{}

\newcommand\scrolls{\textsc{Scrolls}}
\newcommand\zs{Zero\textsc{Scrolls}}
\newcommand\booksum{BookSum}
\newcommand\booksumsort{BookSumSort}

\newif\ifcomments
\commentstrue
\ifcomments
    \def \ifempty#1{\def\temp{#1} \ifx\temp\empty }

\title{ZeroSCROLLS: A Zero-Shot Benchmark for Long Text Understanding}

\author{
Uri Shaham$^\tau$ \quad Maor Ivgi$^\tau$ \quad Avia Efrat$^\tau$ \quad Jonathan Berant$^\tau$ \quad Omer Levy$^{\tau\mu}$ \\
\\
$^\tau$ The Blavatnik School of Computer Science, Tel Aviv University\\
$^\mu$ Meta AI\\
}

\begin{document}
\maketitle

\begin{abstract}
We introduce \zs{}, a zero-shot benchmark for natural language understanding over long texts, which contains only test and small validation sets, without training data. We adapt six tasks from the \scrolls{} benchmark, and add four new datasets, including two novel information fusing tasks, such as aggregating the percentage of positive reviews.
Using \zs{}, we conduct a comprehensive evaluation of both open-source and closed large language models, 
finding that Claude outperforms ChatGPT, and that GPT-4 achieves the highest average score.
However, there is still room for improvement on multiple open challenges in \zs{}, such as aggregation tasks, where models struggle to pass the naive baseline.
As the state of the art is a moving target, we invite researchers to evaluate their ideas on the live \zs{} leaderboard.\footnote{\url{https://www.zero.scrolls-benchmark.com/}}
\end{abstract}

\section{Introduction}

Large language models (LLMs) have been improving at an incredible pace, solving problems that seemed out of reach, without any task-specific training examples \cite{wei2022finetuned, ouyang2022training, openai2023gpt4}. 
As commercial LLMs are adopted worldwide, it becomes clear that they must also operate successfully over long sequences, such as conversation histories or scientific documents.
However, current LLM benchmarks that do evaluate models in a zero-shot setting, such as HELM \citep{liang2022holistic} and BigBench \citep{srivastava2022imitation}, mostly focus on short sequences; BigBench, for example, has an average of 77 words per input.
To fill this gap, we introduce \zs{}: Zero-Shot CompaRison Over Long Language Sequences, a benchmark for zero-shot long text reasoning over natural language, and conduct a thorough investigation of state-of-the-art LLMs.

\begin{figure}[t]
\centering
\includegraphics[width=0.47\textwidth]{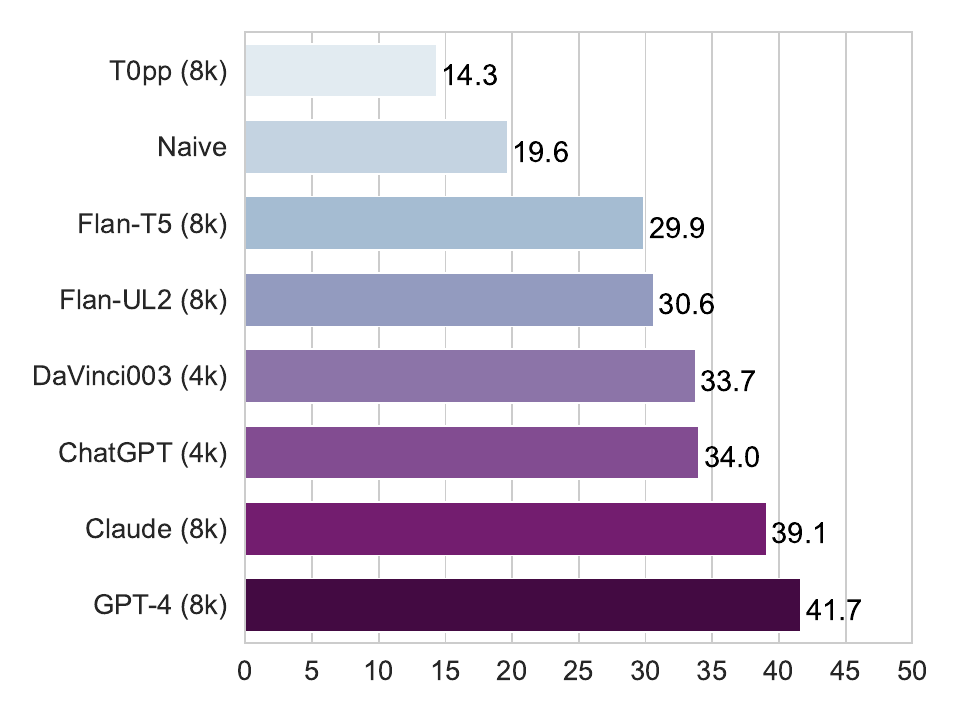}
\caption{\zs{} measures the average performance of state-of-the-art language models across 10 long text understanding tasks. The maximal amount of tokens each model can process is given in parentheses.}
\label{fig:zs_score}
\end{figure}

\begin{figure*}[t]
  \centering
  \includegraphics[width=\textwidth]{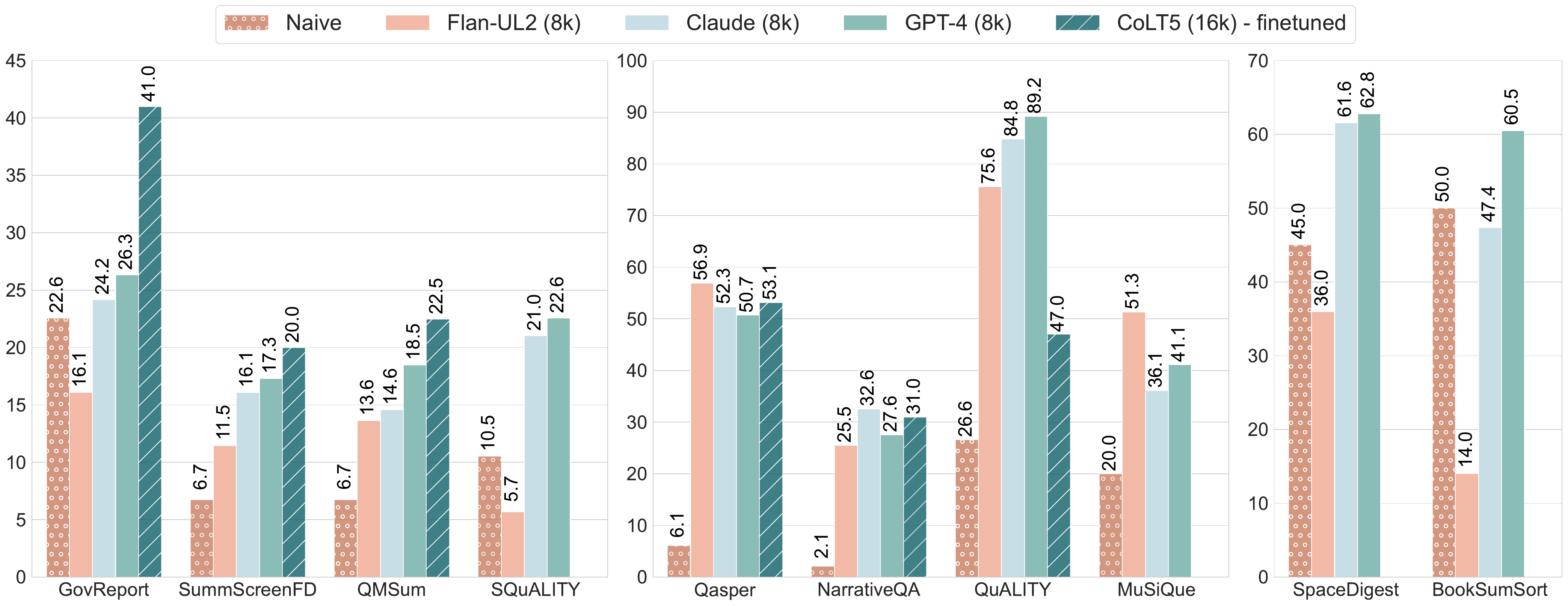}
  \caption{Per task scores of various LLMs and other baselines. In parentheses: the maximum number of tokens.}
\label{fig:bars_two_cols}
\end{figure*}

\zs{} extends \scrolls{} \citep{shaham-etal-2022-scrolls}, a long text understanding benchmark that enables fine-tuning, adding four additional tasks: query-based summarization, multi-hop question answering, sentiment aggregation, and sorting book chapter summaries.
We specifically design the latter two tasks to examine a model's ability to aggregate and compare information across long sequences, while keeping evaluation simple and accurate.
\zs{} is designed to test \textit{zero-shot} capabilities, and contains test sets with simple natural prompts and private gold references, small validation sets, and no train data. 
It has a live leaderboard to enable transparent and dynamic progress.
Figure~\ref{fig:zs_score} shows the state of the leaderboard based on our experiments, and \cref{fig:bars_two_cols} shows a per-task breakdown of a selected subset of models.

We use this new testbed to perform extensive evaluation and analysis across state-of-the-art open and closed models.
On question answering tasks, we find that zero-shot LLMs bridge the gap with task-specific fine-tuned models; GPT-4 sets a new state of the art on the challenging QuALITY task \cite{pang-etal-2022-quality}, almost reaching human performance. 
In contrast, LLMs generally struggle to obtain such high scores for summarization tasks without a training set from which to learn the nuances and artifacts of each dataset, even though GPT-4 does approach the fine-tuned state of the art on two of three datasets.
We also observe that two of our new tasks, sentiment aggregation and sorting book chapter summaries, prove exceptionally challenging for all LLMs, with only GPT-4 surpassing the naive baseline in each task. Our code is available online.\footnote{https://github.com/tau-nlp/zero\_scrolls}

When analyzing GPT-4 responses, we often find correct answers that do not match the requested format; e.g. producing a full sentence when asked to answer in a single phrase.
This problem is not unique to GPT-4, as different models may deviate from the specified format in different tasks.
While \zs{} is primarily aimed at facilitating research in understanding long texts, we encourage the community to use this benchmark to advance research in instruction understanding, prompt engineering, and evaluation of generated texts as well.

\section{Background: \scrolls{}}

\scrolls{} \cite{shaham-etal-2022-scrolls} was introduced as a long text understanding benchmark.
Its datasets were curated, cleaned, and reformatted to a single input-output format allowing for easy and fast usage, with every example containing a single long document, such as a scientific paper or a book.
Since its launch, \scrolls{} has facilitated significant progress, including new pretraining objectives \cite{tay2023ul}, adaptations of short-text models to long sequences \cite{ phang2022investigating, xiong2022adapting,10.1162/tacl_a_00547,bertsch2023unlimiformer}, and dedicated long sequence models pretrained from scratch \cite{guo-etal-2022-longt5, ainslie2023colt5}.

All the aforementioned methods eventually fine-tune a specialized model for every single task in \scrolls{}, a setting that remains important for many applications. 
However, in the modern era of general purpose, zero-shot reasoning LLMs, a new evaluation setup is required, where this dependence on task-specific fine-tuning is alleviated.

\section{The \zs{} Benchmark}
\label{sec:daa}

\zs{} is a zero-shot benchmark containing test sets of ten natural language tasks, each one requiring reasoning over a different type of long text.
To ensure affordability, we limit every task to a maximum of 500 examples.

\subsection{Tasks}

We describe the different \zs{} datasets, six of which we adapt from \citet{shaham-etal-2022-scrolls}, and four new tasks. \cref{tab:datasets_statistics} provides an overview.

\begin{table*}[th]
    \small
    \centering
    \begin{tabular}{@{}llllrr@{}}
    \toprule
    \textbf{Dataset} & \textbf{Task} & \textbf{Domain} & \textbf{Metric} & \textbf{Avg \#Words} &  \textbf{\#Examples}  \\
    \midrule
    GovReport \cite{huang-etal-2021-efficient} & Summarization & Government & ROUGE & 7,273 & 500\\
    SummScreenFD \cite{chen-etal-2022-summscreen} & Summarization & TV & ROUGE & 5,663 & 337 \\
    QMSum \cite{zhong-etal-2021-qmsum} & QB-Summ & Meetings & ROUGE & 10,839  & 281 \\
    SQuALITY \cite{wang-etal-2022-squality} & QB-Summ & Literature & ROUGE & 4,971  & 260 \\
    Qasper \cite{dasigi-etal-2021-dataset} & QA & Science & F1 & 3,531  & 500 \\
    NarrativeQA \cite{kocisky-etal-2018-narrativeqa} & QA & Literature, Film & F1 & 49,384  & 500 \\
    QuALITY \cite{pang-etal-2022-quality} & MC-QA & Literature, Misc & Accuracy & 4,248  & 500 \\
    MuSiQue \cite{trivedi-etal-2022-musique} & QA & Wikipedia & F1 & 1,749  & 500 \\
    SpaceDigest (New) & Aggregation & Reviews & ES & 5,481  & 500 \\
    \booksumsort{} (New) & Aggregation & Literature &  C\textsubscript{idx} & 6,840 & 500 \\
    \bottomrule
    \end{tabular}
    \caption{An overview of the data statistics in \zs{}. \textit{QB-Summ} means query-based summarization, \textit{MC-QA} abbreviates multiple-choice question answering. \textit{ES} refers to exponential similarity and \textit{C\textsubscript{idx}} refers to concordance index. SpaceDigest data is from on the Space dataset \cite{angelidis-etal-2021-extractive} and \booksumsort{} data is from the BookSum dataset \citep{kryscinski-etal-2022-booksum}.}
    \label{tab:datasets_statistics}
    \end{table*}

\subsubsection{Summarization}

We adopt the three summarization datasets from \scrolls{} (GovReport, SummScreenFD, and QMSum), and add a fourth (SQuALITY). GovReport and SummScreenFD are full-document summarization tasks, while QMSum and SQuALITY are query-focused.

\paragraph{GovReport}
\citep{huang-etal-2021-efficient} contains long reports by the Congressional Research Service 
 and the U.S. Government Accountability Offices, with their expert written summaries.

\paragraph{SummScreenFD}
\citep{chen-etal-2022-summscreen} contains episode scripts from TV shows with community contributed recaps that were collected from Wikipedia and TVMaze as their summaries.

\paragraph{QMSum}
\citep{zhong-etal-2021-qmsum} is a query-based summarization dataset over meetings transcripts. 
It contains academic meetings, industrial product meetings, and Welsh and Canadian parliament transcripts.
Alongside the meeting transcript, each instance contains a query, which aims to focus the summary on a particular topic.

\paragraph{SQuALITY}
\citep{wang-etal-2022-squality} is a question-focused summarization dataset, where given a story from Project Gutenberg, the task is to produce a summary of the story or aspects of it based on a guiding question. The questions and summaries are original and crowdsourced; experienced writers were told to design questions that require reading significant parts of the story to answer correctly.

\subsubsection{Question Answering}

We adopt the three question answering datasets from \scrolls{} (Qasper, NarrativeQA, and QuALITY), and add MuSiQue, which focuses on multi-hop question answering.

\paragraph{Qasper}
\cite{dasigi-etal-2021-dataset} contains NLP papers from the Semantic Scholar Open Research Corpus (S2ORC) \cite{lo-etal-2020-s2orc}.
NLP practitioners provided questions based on the abstracts, and another set of practitioners answered given the articles.

\paragraph{NarrativeQA}
\cite{kocisky-etal-2018-narrativeqa} contains questions and answers over books from Project Gutenberg  and movie scripts from various websites. To create questions and answers, annotators were provided summaries of the books and movies from Wikipedia, and each question was answered by one or more annotators.

\paragraph{QuALITY}
\cite{pang-etal-2022-quality} contains stories and articles from Project Gutenberg, the Open American National Corpus, and more. Each instance contains a story and a multiple choice question; question writers were guided to write questions that require reading large portions of the story to answer correctly.

\paragraph{MuSiQue}
\citep{trivedi-etal-2022-musique} is a multi-hop question answering dataset, where the inputs are 20 Wikipedia paragraphs and a question that requires multiple hops between different paragraphs.
In the original dataset, each question also has an unanswerable twin question, where the correct answer is not present in the paragraphs. We randomly sample 100 unanswerable and 400 answerable questions for \zs{}.

\subsubsection{Aggregation}
\label{sec:aggregation}

We create two new tasks that, by construction, require contextualizing and aggregating information from different parts of the input.
Despite the inherent complexity required to solve these tasks, we design their evaluation to be simple and accurate.

\paragraph{SpaceDigest}
is a new sentiment aggregation task.
Given 50 hotel reviews (without their ratings) from the Space dataset \cite{angelidis-etal-2021-extractive}, the task is to determine the percentage of positive reviews.
We create one example (50 reviews) per hotel from the 500 most rated hotels in the original dataset, keeping only strictly positive (rating 5 or 4) or negative (rating 2 or 1) reviews, discarding ones with an ambivalent rating of 3.
To verify that humans perform this task well, we gave 5 human annotators a shortened version of the examples (containing 10 reviews per example) and asked them to write the percentage of positive reviews. Each annotator was assigned 10 examples (100 reviews per annotator, 500 overall). The annotators aggregated their individual predictions perfectly, and had a total of 8 single-review classification errors out of the 500 reviews seen (\(\sim \)98.4\% accuracy).

\paragraph{\booksumsort{}}
is a new task based on the \booksum{} dataset \citep{kryscinski-etal-2022-booksum}, which contains summaries of chapters (or parts) of novels, plays, and long poems from various sources. Given a shuffled list of chapter summaries, the task is to reorder them according to the original order of summaries in \booksum{}.
We create the task by manually selecting the summaries of 125 books from \booksum{}, retaining only high-quality instances.
We manually edit each summary by removing introductions, prefaces, overviews, and so forth, as well as any other information that may indicate the exact position of a summary; for example, \textit{``Chapter 8 begins with Jane describing...''} is replaced with \textit{``This Chapter begins with Jane describing...''} and \textit{``As the play opens, Hippolytus announces...''} becomes \textit{``Hippolytus announces...''}. Each list of summaries contains between 3 and 86 chapter summaries, with a median of 15 and an average of 18.8 chapters per instance. We select 4 random permutations of each list to create 500 instances.

\subsection{Prompting}
\label{sec:prompt}

\zs{} tests the ability to reason over long texts without any explicit training examples (zero-shot).
We thus complement each data instance with an instruction that defines both the task and the desired output format \cite{efrat2020turking}, without in-context demonstrations.
While we invest effort in designing the canonical prompts for \zs{}, the benchmark is open to further zero-shot prompt engineering \cite{radford2019language, schick-schutze-2021-exploiting, schick-schutze-2021-shot}, such as prompts that encourage chain-of-thought reasoning \cite{NEURIPS2022_9d560961}. \cref{tab:summ_prompts} contains the prompts for the summarization tasks and \cref{tab:qa_and_agg_prompts} contains prompts for question answering and agregation tasks.

\paragraph{Prompt Structure}
\cref{fig:zs_qmsum_example} illustrates an example from the benchmark. 
We manually craft a prompt for each dataset, following a generic template composed of \textit{instruction}, \textit{context}, \textit{query}, and \textit{response}.
The instruction describes the task, and ends with the desired output format (e.g. ``Answer the query in one or more sentences.'' for QMSum).
When the total input size is too long for a model's context window, we trim the context and append a string explicitly stating that the rest of the context is trimmed, to inform the model that it cannot see the entire context.
We then concatenate the context with a header describing what kind of context it is, e.g. ``Report:'', ``Reviews:'', etc.
For tasks that have queries, we append the question or query with an appropriate header.
The prompt ends with a header indicating the response type (e.g. "Answer:" or "Summary:").

\begin{figure}[t]
\small
\centering
{\setlength{\extrarowheight}{3pt}%
\begin{tabular}{|p{0.94\columnwidth}|} 
\hline
\textcolor[HTML]{3078BE}{You are given a meeting transcript and a query
containing a question or instruction. Answer the query in one or more sentences.}\\\\

\textcolor[HTML]{3078BE}{Transcript:}\\

\textcolor[HTML]{000000}{User Interface: That's the same as uh on the top of it uh with the the round uh button.}\\
\textcolor[HTML]{000000}{Industrial Designer: Like this one.}\\

\textcolor[HTML]{000000}{User Interface: But uh we don't uh we don't uh {disfmarker} we do think it's um well {disfmarker} what if with ease of use, w which prefers the {disfmarker} which the the customer of the user prefers.}\\

\textcolor[HTML]{000000}{Industrial Designer: It's important. Uh I think th this is device which which has a learning curve}\textcolor[HTML]{C07830}{... [The rest of the transcript is omitted]}\\\\

\textcolor[HTML]{3078BE}{Query:}\\
\textcolor[HTML]{000000}{What did the group discuss about production costs of the product?}\\\\

\textcolor[HTML]{3078BE}{Answer:}
\vspace{2pt}
\\
\hline
\end{tabular}}
\caption{An example input in \zs{}, taken from the QMSum dataset. The meeting transcript and the question are in \textit{black}, and the \zs{} prompt is in \textit{\textcolor[HTML]{3078BE}{blue}}. In \textit{\textcolor[HTML]{C07830}{copper}} is a string we append to the trimmed context when the model's context window is too short to contain the entire input. }
\label{fig:zs_qmsum_example}
\end{figure}

\paragraph{Accommodations for ChatBots}
Chat LLMs, such as ChatGPT and Claude, are designed to interact with humans through a chat interface.
We therefore adapt our canonical prompts to accommodate these models.
Specifically, omit the response header (e.g. ``Summary:'' or ``Answer:'') as it is clear, in dialogue, that the input sequence has ended.
In addition, we append ``Do not provide any explanation.'' to the instructions of question answering and aggregation tasks.
For Claude, we wrap each prompt with ``Human:'' and ``Assistant:'' dialogue indicators, and for the question answering and aggregation tasks also add the instruction to ``please highlight your final answer with <\{response\_type\}></\{response\_type\}> tags'' -- as recommended by Anthropic's documentation.\footnote{\url{https://console.anthropic.com/docs/prompt-design/classification}}

\subsection{Automatic Evaluation}
\label{subsec:metrics}

\zs{} evaluation is fully automatic.
Given a model's response to every test instance, we apply per-task automatic evaluation metrics.
These are then averaged across tasks to produce the model's \zs{} score.
For existing datasets, we follow \citet{shaham-etal-2022-scrolls} and use the metrics provided by each dataset's authors.
For our newly proposed tasks (SpaceDigest and \booksumsort{}), we use two new automatic metrics.

\paragraph{ROUGE}
\textit{(GovReport, SummScreenFD, QMSum, SQuALITY)}
ROUGE \cite{lin-2004-rouge} measures ngram overlap between generated and reference summaries. For each instances, we combine ROUGE-1, ROUGE-2, and ROUGE-L into a single score by computing their geometric mean.
For SQuALITY, where there are multiple references, we take the maximal value of each ROUGE type before computing the geometric mean.

\paragraph{F1}
\textit{(Qasper, NarrativeQA, MuSiQue)}
F1 computes unigram overlap between generated and reference answers, after normalizing white-spaces, lowercasing, omitting stopwords and punctuation \cite{rajpurkar-etal-2016-squad}, and transliterating any Unicode text to ASCII characters.
For Qasper and NarrativeQA, where there are multiple reference answers, we take the maximal F1 score per instance.

\paragraph{Accuracy}
\textit{(QuALITY)}
For multiple choice questions, we compare the predicted letter (A, B, C, or D) to the reference. We use the first valid option letter surrounded by word boundaries.
    
\paragraph{Exponential Similarity}
\textit{(SpaceDigest)}
Assuming that the output is a percentage,\footnote{If the output is not a percentage, we score 0\%. We parse the first appearance of a percentage; e.g. for the output \textit{``Out of 50 reviews, 20 are positive and 30 are negative, so 40\% of the reviews are positive 60\% are negative.''} we automatically parse 40\% as the answer.}
we compute the exponential similarity between the gold reference percentage $p$ and the predicted scalar $\hat{p}$: 
$$ ES(p, \hat{p})= d^{-c \cdot |p - \hat{p}|} $$
We use $d = 2$ and $c = 10$, which means that, intuitively, the score gets cut by half for every 10 point deviation from the correct answer.

\paragraph{Concordance Index}
\textit{(\booksumsort{})}
Assuming that the output is a permutation of the given chapter summary IDs,\footnote{If the output is not a permutation, we score 0\%. We discard all characters but digits, commas, and white-spaces from the output string to eliminate any prefixes such as ``Order:''}
we measure the amount of chapter summary pairs that are in the right order, divided by the total number of pairs $\binom{n}{2}$.
The average random permutation scores 50\% on this metric.

\section{Evaluating State-of-the-Art LLMs}

Using \zs{} we conduct, to the best of our knowledge, the first systematic LLMs zero-shot performance comparison over tasks that require long text understanding.

\subsection{Models}

We evaluate both open-source models and closed products available via APIs.
We apply greedy decoding to all models, and leave further research into other decoding strategies to future work.
Table~\ref{tab:models_details} shows the selection of models we evaluate.

\begin{table}[t]
    \small
    \centering
    \begin{tabular}{@{}lccc@{}}
    \toprule
   \multirow{2}{*}{\textbf{Model}} & \multirow{2}{*}{\textbf{Params}} & \multicolumn{1}{@{}c}{\textbf{Maximum}} &  \multirow{2}{*}{\textbf{Open/Closed}}    \\
        &  & \multicolumn{1}{@{}c}{\textbf{Length}} &  \\
    \midrule
    T0pp & 11B & 8,192 & Open   \\
    Flan-T5 & 11B & 8,192 & Open  \\
    Flan-UL2 & 20B & 8,192 & Open   \\
    DaVinci003 & -- & 4,096 & Closed   \\
    ChatGPT & -- & 4,096 & Closed   \\
    Claude & -- & 8,192 & Closed   \\
    GPT-4 & -- & 8,192 & Closed   \\
    \bottomrule
    \end{tabular}
    \caption{State of the art LLMs we evaluate. Exact parameter counts of closed models are not publicly available.}
    \label{tab:models_details}
\end{table}

\begin{table*}[th]
\small
\centering
\begin{tabular}{@{}l@{}rccccccccccc@{}}
\toprule
\multirow{2}{*}{\textbf{Model}} & \multirow{2}{*}{\textbf{Tokens}} & \textbf{GvRp} & \textbf{SSFD} &  \textbf{QMsm} & \textbf{SQAL}  & \textbf{Qspr} & \textbf{Nrtv} &  \textbf{QALT} & \textbf{MuSQ} &  \textbf{SpDg} & \textbf{BkSS}  & \multirow{2}{*}{\textbf{Avg}} \\
&  & R\textsubscript{geo}  &  R\textsubscript{geo}  &  R\textsubscript{geo}  & R\textsubscript{geo} & F1  &  F1  &  AC  & F1 &  ES  & C\textsubscript{idx} &  \\
\midrule
\multicolumn{13}{@{}l}{\textbf{\textit{Baselines}}}\\
Naive & - & 22.6 & 6.7 & 6.7 & 10.5 & 6.1 & 2.1 & 26.6 & 20.0 & 45.0 & 50.0 & \textit{19.6}\\
Human & - & - & - & - & 23.6 & 67.7 & 58.2 & 93.5 & 74.8 & 93.3 & - & -\\
\midrule
\multicolumn{13}{@{}l}{\textbf{\textit{Open Source Models}}}\\
T0pp & 8192 & 7.1 & 9.6 & 7.2 & 3.9 & 25.0 & 18.7 & 21.4 & 35.3 & 15.2 & 0.0 & \textit{14.3}\\
Flan-T5 & 8192 & 17.6 & 7.8 & 11.0 & 8.0 & 48.3 & 19.3 & 75.2 & 46.8 & 48.7 & 16.4 & \textit{29.9}\\
Flan-UL2 & 8192 & 16.1 & 11.5 & 13.6 & 5.7 & \textbf{56.9} & 25.5 & 75.6 & \textbf{51.3} & 36.0 & 14.0 & \textit{30.6}\\
\midrule
\multicolumn{13}{@{}l}{\textbf{\textit{Closed Models}}}\\
DaVinci003 & 4096 & 21.7 & 16.1 & 16.9 & 22.0 & 52.7 & 24.6 & 69.0 & 33.5 & 31.3 & 49.5 & \textit{33.7}\\
ChatGPT & 4096 & 21.3 & 16.1 & 15.6 & 20.4 & 49.3 & 25.1 & 66.6 & 27.1 & 49.1 & 49.8 & \textit{34.0}\\
Claude & 8000 & 24.2 & 16.1 & 14.6 & 21.0 & 52.3 & \textbf{32.6} & 84.8 & 36.1 & 61.6 & 47.4 & \textit{39.1}\\
GPT-4 & 8192 & \textbf{26.3} & \textbf{17.3} & \textbf{18.5} & \textbf{22.6} & 50.7 & 27.6 & \textbf{89.2} & 41.1 & \textbf{62.8} & \textbf{60.5} & \textit{\textbf{41.7}}\\
\midrule
\multicolumn{13}{@{}l}{\textbf{\textit{Fine-tuned Models}}}\\
CoLT5 & 16384 & 41.0 & 20.0 & 22.5 & - & 53.1 & 31.0 & 47.0 & - & - & - & -\\
\bottomrule
\end{tabular}
\caption{The \zs{} leaderboard, at the time of writing. The dataset abbreviations stand for: GovReport, SummScreenFD, QMSum, SQuALITY, Qasper, NarrativeQA, QuALITY, MuSiQue, SpaceDigest, \booksumsort{}.}
\label{tab:results}
\end{table*}

\paragraph{Open Source Models}
We experiment with \textbf{Flan-T5-xxl} \citep{wei2022finetuned} and \textbf{Flan-UL2}, the instruction-tuned versions of T5 \citep{JMLR:v21:20-074} and UL2 \citep{tay2023ul}, as well as \textbf{T0pp} \citep{sanh2022multitask}, an LM-adapted \cite{lester-etal-2021-power} version of T5 that was finetuned on various NLP tasks for zero shot generalization.
For all open-source models we use a maximum input length of 8,192 tokens (larger contexts were unstable).
We also experiment with shorter context lengths and smaller variants of Flan-T5.

\paragraph{Closed Models (Products)}
Using product APIs, we evaluate \textbf{Claude} v1.3 from Anthropic,\footnote{https://www.anthropic.com/index/introducing-claude} and  \textbf{DaVinci003},\footnote{https://platform.openai.com/docs/model-index-for-researchers}  \textbf{ChatGPT} v0301,\footnote{https://chat.openai.com/} and \textbf{GPT-4} v0314 \citep{openai2023gpt4} from OpenAI.
The maximal context length of these models includes both input and output.

\paragraph{Task-Specific Models}
To compare general-purpose LLMs (zero-shot) to task-specific models (fine-tuned), we use predictions by \textbf{CoLT5-xl} \citep{ainslie2023colt5}, a transformer allocating more resources to important tokens, with a maximum input length of 16,384 tokens and is the current state of the art on \scrolls{}.

\paragraph{Naive Baselines}
We implement simple baselines for all tasks.
For GovReport, SummScreenFD, QMSum, SQuALITY and NarrativeQA, we select random spans from the input document of 500, 200, 50, 120 and 4 words respectively.
For Qasper, we randomly decide whether to use one of its fixed choices (``Yes'', ``No'', ``Unanswerable'') or choose a random span of 15 words.
For MuSiQue, we use ``Unanswerable'' for every instance.
For QuALITY, we randomly select an option from A, B, C, or D.
For SpaceDigest we always use 50\%, and for \booksumsort{} we use the trivial permutation ``$1, 2, 3,..., n$.''

\paragraph{Human Performance}
We provide human performance figures for 6 of the 10 tasks.
For SQuALITY, \citet{wang-etal-2022-squality} estimate human performance by comparing one reference against the other three.
Similarly, for Qasper and NarrativeQA, we calculate inter-annotator F1 on the \zs{} subsets.
We use the human scores reported by \citet{pang-etal-2022-quality} on the full QuALITY test set,
while for MuSiQue, we combine statistics on answerable and non-answerable sets from \citet{trivedi-etal-2022-musique}.
For SpaceDigest, we use our own human annotations (\cref{sec:aggregation}) to estimate exponential similarity over 50 reviews.

\subsection{Main Results}

\cref{tab:results} shows the results for every model on every \zs{} task, along with the average.
The overall best model is GPT-4 with an average score of 41.7, and its closest competitor is Claude with 39.1, both significantly higher than the other models.
We discuss the results per task category.

\paragraph{Summarization}
There is a clear trend where the open-source models lag behind product-grade LLMs, and that GPT-4 reaches the highest ROUGE scores on all four datasets. However, zero-shot LLMs struggle to compete with models fine-tuned per dataset (CoLT5) on those tasks, with some gap on SummScreenFd and QMSum, and a dramatic difference on GovReport (41.0 compared to 26.3). In SQuALITY, GPT-4 is only one point away from the lower bound on human performance.

\begin{table*}[t]
\small
\centering
\begin{tabular}{@{}l@{}rccccccccccc@{}}
\toprule
\multirow{2}{*}{\textbf{Model}} & \multirow{2}{*}{\textbf{Tokens}} & \textbf{GvRp} & \textbf{SSFD} &  \textbf{QMsm} & \textbf{SQAL}  & \textbf{Qspr} & \textbf{Nrtv} &  \textbf{QALT} & \textbf{MuSQ} &  \textbf{SpDg} & \textbf{BkSS}  & \multirow{2}{*}{\textbf{Avg}} \\
&  & R\textsubscript{geo}  &  R\textsubscript{geo}  &  R\textsubscript{geo}  & R\textsubscript{geo} & F1  &  F1  &  AC  & F1 &  ES  & C\textsubscript{idx} &  \\
\midrule
\multicolumn{13}{@{}l}{\textbf{\textit{Flan-T5 Across Model Sizes}}}\\
Flan-T5-s & 8192 & 7.6 & 4.2 & 8.3 & 3.8 & 18.5 & 11.6 & 34.6 & 21.0 & 0.0 & 0.0 & \textit{11.0}\\
Flan-T5-b & 8192 & 5.4 & 5.1 & 9.7 & 5.6 & 14.2 & 16.5 & 48.4 & 26.9 & 0.0 & 0.3 & \textit{13.2}\\
Flan-T5-l & 8192 & 6.9 & 6.8 & 9.7 & 5.7 & 33.6 & 20.1 & 62.4 & 33.1 & 48.0 & 0.3 & \textit{22.7}\\
Flan-T5-xl & 8192 & 15.2 & 7.2 & 10.2 & 6.6 & 46.6 & \textbf{21.6} & 69.6 & 42.8 & 32.8 & 2.2 & \textit{25.5}\\
Flan-T5-xxl & 8192 & \textbf{17.6} & \textbf{7.8} & \textbf{11.0} & \textbf{8.0} & \textbf{48.3} & 19.3 & \textbf{75.2} & \textbf{46.8} & \textbf{48.7} & \textbf{16.4} & \textit{\textbf{29.9}}\\
\midrule
\multicolumn{13}{@{}l}{\textbf{\textit{Flan-T5-xxl Across Input Lengths}}}\\
Flan-T5-xxl & 512 & 10.0 & 7.9 & 10.4 & 6.1 & 15.3 & 17.6 & 48.2 & 26.0 & 20.8 & 9.0 & \textit{17.1}\\
Flan-T5-xxl & 1024 & 12.1 & 9.4 & 10.1 & 6.3 & 25.5 & 18.9 & 53.2 & 30.3 & 28.7 & 13.4 & \textit{20.8}\\
Flan-T5-xxl & 2048 & 14.0 & \textbf{10.0} & 11.0 & 6.8 & 35.7 & 20.9 & 59.8 & 40.6 & 35.0 & 14.7 & \textit{24.9}\\
Flan-T5-xxl & 4096 & 17.3 & 9.1 & \textbf{11.8} & 7.4 & 46.5 & \textbf{22.2} & 70.8 & \textbf{46.8} & 44.1 & 15.1 & \textit{29.1}\\
Flan-T5-xxl & 8192 & \textbf{17.6} & 7.8 & 11.0 & \textbf{8.0} & \textbf{48.3} & 19.3 & \textbf{75.2} & \textbf{46.8} & \textbf{48.7} & \textbf{16.4} & \textit{\textbf{29.9}}\\
\midrule
\multicolumn{13}{@{}l}{\textbf{\textit{Claude Across Input Lengths}}}\\
Claude & 4096 & 23.0 & 15.0 & 14.3 & 20.2 & 47.7 & 31.7 & 76.8 & 35.8 & 61.1 & 37.6 & \textit{36.3}\\
Claude & 8000 & \textbf{24.2} & \textbf{16.1} & \textbf{14.6} & \textbf{21.0} & \textbf{52.3} & \textbf{32.6} & \textbf{84.8} & \textbf{36.1} & \textbf{61.6} & \textbf{47.4} & \textit{\textbf{39.1}}\\
\bottomrule
\end{tabular}
\caption{Performance of Flan-T5 across model sizes, and Flan-T5 and Claude across input lengths.}
\label{tab:size_and_length}
\end{table*}

\paragraph{Question Answering}
We see a different trend in question answering. GPT-4 achieves the best result on only one dataset, QuALITY, where it scores 89.2, close to human performance of 93.5. Flan-UL2 sets the high scores for Qasper and MuSiQue, while Claude has the best F1 on NarrativeQA, 5 points more than GPT-4. Our analysis in \cref{sec:analysis} reveals that GPT-4 does not conform to the required answer format, resulting in a lower score.

\paragraph{Aggregation}
Our new SpaceDigest and \booksumsort{} datasets enrich \zs{} with challenges that explicitly require aggregating information across the sequence.
Results indicate that both tasks are difficult for current LLMs.
Performance figures for SpaceDigest show that even though sentiment analysis, counting, and divisions are all ``easy'' tasks for contemporary models, their combination can be quite challenging; only Claude and GPT-4 significantly outperform the naive baseline.
The situation is even more dire in \booksumsort{}, where only GPT-4 outperforms the naive baseline.

\subsection{Impact of Model Size and Input Length}

We now discuss the effects of increasing model size (parameters) and context length (tokens).
As one may expect, both dimensions improve performance on \zs{}, suggesting that the benchmark does indeed necessitate complex reasoning over long sequences.

\paragraph{Model Size}
The upper section of \cref{tab:size_and_length} shows results of Flan-T5 of various sizes, ranging from S (60M parameters) to XXL (11B parameters). As expected, increasing model size drives performance upwards across almost all tasks.

\paragraph{Input Length}
The middle and lower sections of \cref{tab:size_and_length} show the effect of increasing the maximum number of input tokens for Flan-T5 and Claude. In general, increasing the number of tokens helps the models preform the tasks better. Claude is able to utilize the extra tokens more consistently, which results in an almost 3 point increment to its average score when going from 4k to 8k tokens. Interestingly, Flan-T5 also achieves higher scores on longer inputs in many cases, despite being trained on much shorter sequences.

\section{Analysis}
\label{sec:analysis}

While GPT4 has the highest score on the \zs{} leaderboard, we find it surprising that other models score higher on a number of question answering tasks.
We analyze model generations and observe that GPT-4 responses do not match the desired output format (despite explicit instructions in the prompt), which results in penalization by the automatic metrics.
Further analysis reveals that format discrepancy is a phenomenon that occurs across different LLMs and tasks, and is not unique to GPT-4 and question answering.

\begin{figure}[t]
\centering
   \includegraphics[width=1\linewidth]{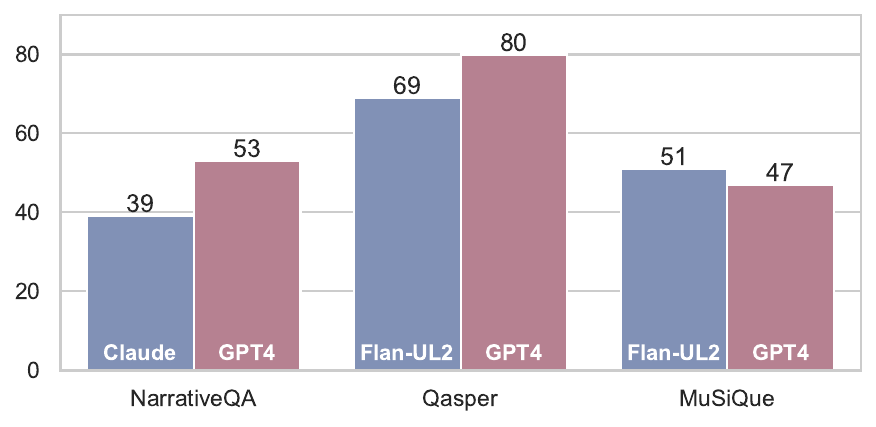}
  \caption{Human evaluation (accuracy) over 100 questions from NarrativeQA, Qasper, and MuSiQue, comparing GPT-4 to the highest scoring model of each dataset.}
\label{fig:correct} 
\end{figure}

\paragraph{Discrepancies in Question Answering}
\label{sec:disc_qa}
We analyze the responses of GPT-4 and Claude for NarrativeQA (where Claude scores 5 points higher), and the responses of GPT-4 and Flan-UL2 for Qasper and MuSiQue (where Flan-UL2 scores 6.2 and 10.2 points higher, respectively).
Specifically, we sample 100 instances from each dataset, and annotate whether the answer is correct, ignoring formatting, fluency, or other factors.
\cref{fig:correct} shows that, in contrast to the F1 scores, GPT-4 performs better than Claude and Flan-UL2 on NarrativeQA and Qasper, respectively, and that the gap between GPT-4 and Flan-UL2 on MuSiQue is smaller in practice.

From examining the generated texts, we learn that GPT-4 consistently generates complete answers even though the prompt instructs otherwise (see Section~\ref{sec:prompt} and Appendix~\ref{sec:appendix}).
We further analyze 200 random instances from NarrativeQA and check whether GPT-4 and Claude respond in the specified format, i.e. ``using a single phrase if possible,'' regardless of whether the content is correct or not.
While Claude answers 191 questions in the correct format, GPT-4 does so for only 71 out of the 200 analyzed examples -- explaining why GPT-4 is penalized harder by the F1 metric, despite being ``correct'' more often than Claude.\footnote{Another interesting observation from analyzing NarrativeQA is that GPT-4 sometimes responds that it is unable to answer the question because the (trimmed) context does not contain the answer. It does so for 30 out of 200 cases, while Claude generates a similar response for only 5, despite both models having similar context lengths (8k).}

\begin{figure}[t!]
\centering
   \includegraphics[width=0.85\linewidth]{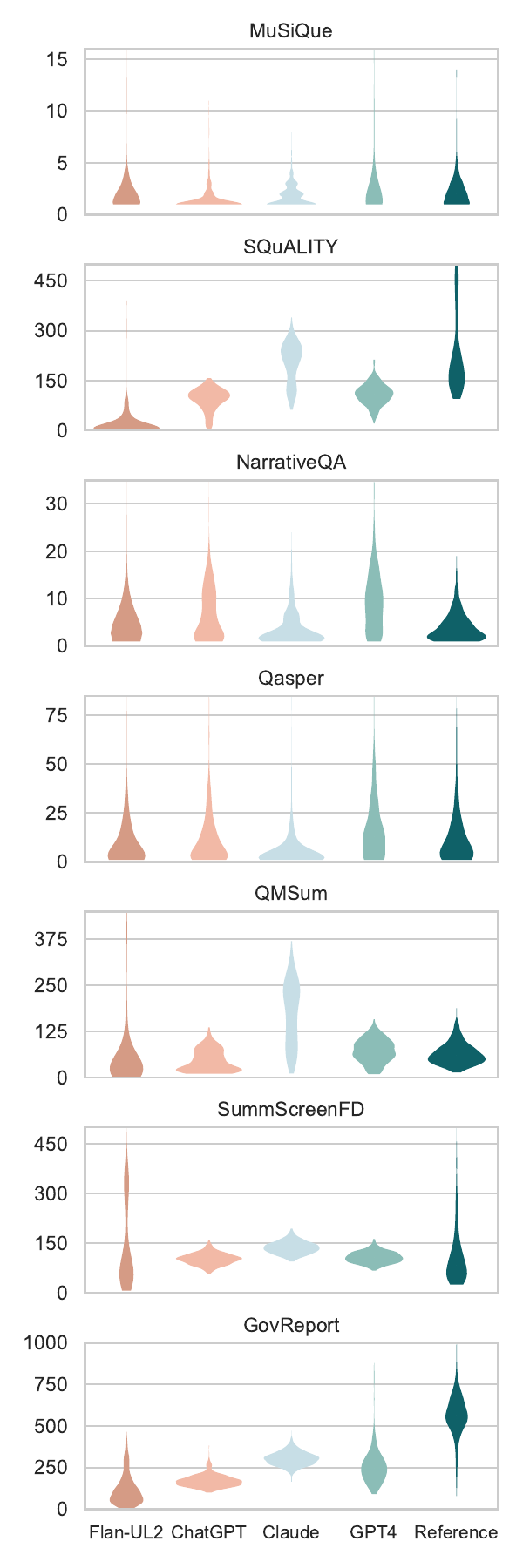}
  \caption{Distribution of the number of generated words.}
\label{fig:output_len} 
\end{figure}

\paragraph{Format Discrepancy}
\Cref{fig:output_len} surveys the distribution of output lengths across multiple tasks and models.
In most cases, models generate outputs that fall within the distribution of reference lengths, indicating that the format criteria provided in the prompts are sufficient.
However, certain task-model combinations fall outside of the reference distribution.
While the NarrativeQA plot confirms our previous observation that GPT-4 generates longer answers for this task, we find that format discrepancy is not unique to this dataset or GPT-4, as different models struggle to generate texts in the correct format on different tasks;
Claude generates long answers for QMSum, Flan-UL2 generates long summaries in SummScreenFD, and all models generate short summaries for GovReport, which negatively impacts their scores. 

\section{Conclusion}
We introduce \zs{}, a benchmark for zero-shot natural language understanding over long texts.
\zs{} enables systematic comparison of LLMs on tasks with naturally long input texts, and ones that require contextualizing and aggregating information from multiple documents.
We evaluate open-source and production-grade LLMs to find that GPT-4 and Claude are currently the best performing models, while open-source models such as Flan-UL2 also prove powerful at long-context question answering tasks.
\zs{} remains an open challenge for LLM research, with our two new aggregation tasks proving to be particularly difficult for contemporary LLMs. 

\section{Limitations}

As language models improve, evaluating them presents a growing challenge given their ability to consistently generate coherent and reasonable text, which is harder to score, even with gold references at hand.
Specifically in the zero-shot setting, where models must infer the output format from the prompt, ROUGE and F1 (ngram metrics) can assign low scores for semantically equivalent generations, with different word choices or answer lengths.
Additionally, to conduct fair evaluation, we use common prompt templates across models for every task, while model-specific prompts, as well as chain-of-thought prompting may improve model performance on this benchmark.
Finally, the state of the art is a moving target, and as we write these lines new long-range models, alignment methods, decoding algorithms, and prompting techniques become available; we invite researchers to evaluate their ideas on the \zs{} leaderboard.

\section*{Acknowledgements}
This research is supported by the Yandex Initiative in Machine Learning and by the Len Blavatnik and the Blavatnik Family foundation. The benchmark is released by Tel Aviv University. All experiments were conducted by Tel Aviv University.

\bibliographystyle{acl_natbib}
\bibliography{anthology,custom}

\begin{thebibliography}{34}
\expandafter\ifx\csname natexlab\endcsname\relax\def\natexlab#1{#1}\fi

\bibitem[{Ainslie et~al.(2023)Ainslie, Lei, de~Jong, Ontañón, Brahma,
  Zemlyanskiy, Uthus, Guo, Lee-Thorp, Tay, Sung, and
  Sanghai}]{ainslie2023colt5}
Joshua Ainslie, Tao Lei, Michiel de~Jong, Santiago Ontañón, Siddhartha
  Brahma, Yury Zemlyanskiy, David Uthus, Mandy Guo, James Lee-Thorp, Yi~Tay,
  Yun-Hsuan Sung, and Sumit Sanghai. 2023.
\newblock \href {http://arxiv.org/abs/2303.09752} {Colt5: Faster long-range
  transformers with conditional computation}.

\bibitem[{Angelidis et~al.(2021)Angelidis, Amplayo, Suhara, Wang, and
  Lapata}]{angelidis-etal-2021-extractive}
Stefanos Angelidis, Reinald~Kim Amplayo, Yoshihiko Suhara, Xiaolan Wang, and
  Mirella Lapata. 2021.
\newblock \href {https://doi.org/10.1162/tacl_a_00366} {Extractive opinion
  summarization in quantized transformer spaces}.
\newblock \emph{Transactions of the Association for Computational Linguistics},
  9:277--293.

\bibitem[{Bertsch et~al.(2023)Bertsch, Alon, Neubig, and
  Gormley}]{bertsch2023unlimiformer}
Amanda Bertsch, Uri Alon, Graham Neubig, and Matthew~R. Gormley. 2023.
\newblock \href {http://arxiv.org/abs/2305.01625} {Unlimiformer: Long-range
  transformers with unlimited length input}.

\bibitem[{Chen et~al.(2022)Chen, Chu, Wiseman, and
  Gimpel}]{chen-etal-2022-summscreen}
Mingda Chen, Zewei Chu, Sam Wiseman, and Kevin Gimpel. 2022.
\newblock \href {https://doi.org/10.18653/v1/2022.acl-long.589}
  {{S}umm{S}creen: A dataset for abstractive screenplay summarization}.
\newblock In \emph{Proceedings of the 60th Annual Meeting of the Association
  for Computational Linguistics (Volume 1: Long Papers)}, pages 8602--8615,
  Dublin, Ireland. Association for Computational Linguistics.

\bibitem[{Dasigi et~al.(2021)Dasigi, Lo, Beltagy, Cohan, Smith, and
  Gardner}]{dasigi-etal-2021-dataset}
Pradeep Dasigi, Kyle Lo, Iz~Beltagy, Arman Cohan, Noah~A. Smith, and Matt
  Gardner. 2021.
\newblock \href {https://doi.org/10.18653/v1/2021.naacl-main.365} {A dataset of
  information-seeking questions and answers anchored in research papers}.
\newblock In \emph{Proceedings of the 2021 Conference of the North American
  Chapter of the Association for Computational Linguistics: Human Language
  Technologies}, pages 4599--4610, Online. Association for Computational
  Linguistics.

\bibitem[{Efrat and Levy(2020)}]{efrat2020turking}
Avia Efrat and Omer Levy. 2020.
\newblock \href {http://arxiv.org/abs/2010.11982} {The turking test: Can
  language models understand instructions?}

\bibitem[{Guo et~al.(2022)Guo, Ainslie, Uthus, Ontanon, Ni, Sung, and
  Yang}]{guo-etal-2022-longt5}
Mandy Guo, Joshua Ainslie, David Uthus, Santiago Ontanon, Jianmo Ni, Yun-Hsuan
  Sung, and Yinfei Yang. 2022.
\newblock \href {https://doi.org/10.18653/v1/2022.findings-naacl.55}
  {{L}ong{T}5: {E}fficient text-to-text transformer for long sequences}.
\newblock In \emph{Findings of the Association for Computational Linguistics:
  NAACL 2022}, pages 724--736, Seattle, United States. Association for
  Computational Linguistics.

\bibitem[{Huang et~al.(2021)Huang, Cao, Parulian, Ji, and
  Wang}]{huang-etal-2021-efficient}
Luyang Huang, Shuyang Cao, Nikolaus Parulian, Heng Ji, and Lu~Wang. 2021.
\newblock \href {https://doi.org/10.18653/v1/2021.naacl-main.112} {Efficient
  attentions for long document summarization}.
\newblock In \emph{Proceedings of the 2021 Conference of the North American
  Chapter of the Association for Computational Linguistics: Human Language
  Technologies}, pages 1419--1436, Online. Association for Computational
  Linguistics.

\bibitem[{Ivgi et~al.(2023)Ivgi, Shaham, and Berant}]{10.1162/tacl_a_00547}
Maor Ivgi, Uri Shaham, and Jonathan Berant. 2023.
\newblock \href {https://doi.org/10.1162/tacl_a_00547} {{Efficient Long-Text
  Understanding with Short-Text Models}}.
\newblock \emph{Transactions of the Association for Computational Linguistics},
  11:284--299.

\bibitem[{Ko{\v{c}}isk{\'y} et~al.(2018)Ko{\v{c}}isk{\'y}, Schwarz, Blunsom,
  Dyer, Hermann, Melis, and Grefenstette}]{kocisky-etal-2018-narrativeqa}
Tom{\'a}{\v{s}} Ko{\v{c}}isk{\'y}, Jonathan Schwarz, Phil Blunsom, Chris Dyer,
  Karl~Moritz Hermann, G{\'a}bor Melis, and Edward Grefenstette. 2018.
\newblock \href {https://doi.org/10.1162/tacl_a_00023} {The {N}arrative{QA}
  reading comprehension challenge}.
\newblock \emph{Transactions of the Association for Computational Linguistics},
  6:317--328.

\bibitem[{Kryscinski et~al.(2022)Kryscinski, Rajani, Agarwal, Xiong, and
  Radev}]{kryscinski-etal-2022-booksum}
Wojciech Kryscinski, Nazneen Rajani, Divyansh Agarwal, Caiming Xiong, and
  Dragomir Radev. 2022.
\newblock \href {https://aclanthology.org/2022.findings-emnlp.488} {{BOOKSUM}:
  A collection of datasets for long-form narrative summarization}.
\newblock In \emph{Findings of the Association for Computational Linguistics:
  EMNLP 2022}, pages 6536--6558, Abu Dhabi, United Arab Emirates. Association
  for Computational Linguistics.

\bibitem[{Lester et~al.(2021)Lester, Al-Rfou, and
  Constant}]{lester-etal-2021-power}
Brian Lester, Rami Al-Rfou, and Noah Constant. 2021.
\newblock \href {https://doi.org/10.18653/v1/2021.emnlp-main.243} {The power of
  scale for parameter-efficient prompt tuning}.
\newblock In \emph{Proceedings of the 2021 Conference on Empirical Methods in
  Natural Language Processing}, pages 3045--3059, Online and Punta Cana,
  Dominican Republic. Association for Computational Linguistics.

\bibitem[{Liang et~al.(2022)Liang, Bommasani, Lee, Tsipras, Soylu, Yasunaga,
  Zhang, Narayanan, Wu, Kumar, Newman, Yuan, Yan, Zhang, Cosgrove, Manning,
  Ré, Acosta-Navas, Hudson, Zelikman, Durmus, Ladhak, Rong, Ren, Yao, Wang,
  Santhanam, Orr, Zheng, Yuksekgonul, Suzgun, Kim, Guha, Chatterji, Khattab,
  Henderson, Huang, Chi, Xie, Santurkar, Ganguli, Hashimoto, Icard, Zhang,
  Chaudhary, Wang, Li, Mai, Zhang, and Koreeda}]{liang2022holistic}
Percy Liang, Rishi Bommasani, Tony Lee, Dimitris Tsipras, Dilara Soylu,
  Michihiro Yasunaga, Yian Zhang, Deepak Narayanan, Yuhuai Wu, Ananya Kumar,
  Benjamin Newman, Binhang Yuan, Bobby Yan, Ce~Zhang, Christian Cosgrove,
  Christopher~D. Manning, Christopher Ré, Diana Acosta-Navas, Drew~A. Hudson,
  Eric Zelikman, Esin Durmus, Faisal Ladhak, Frieda Rong, Hongyu Ren, Huaxiu
  Yao, Jue Wang, Keshav Santhanam, Laurel Orr, Lucia Zheng, Mert Yuksekgonul,
  Mirac Suzgun, Nathan Kim, Neel Guha, Niladri Chatterji, Omar Khattab, Peter
  Henderson, Qian Huang, Ryan Chi, Sang~Michael Xie, Shibani Santurkar, Surya
  Ganguli, Tatsunori Hashimoto, Thomas Icard, Tianyi Zhang, Vishrav Chaudhary,
  William Wang, Xuechen Li, Yifan Mai, Yuhui Zhang, and Yuta Koreeda. 2022.
\newblock \href {http://arxiv.org/abs/2211.09110} {Holistic evaluation of
  language models}.

\bibitem[{Lin(2004)}]{lin-2004-rouge}
Chin-Yew Lin. 2004.
\newblock \href {https://aclanthology.org/W04-1013} {{ROUGE}: A package for
  automatic evaluation of summaries}.
\newblock In \emph{Text Summarization Branches Out}, pages 74--81, Barcelona,
  Spain. Association for Computational Linguistics.

\bibitem[{Lo et~al.(2020)Lo, Wang, Neumann, Kinney, and
  Weld}]{lo-etal-2020-s2orc}
Kyle Lo, Lucy~Lu Wang, Mark Neumann, Rodney Kinney, and Daniel Weld. 2020.
\newblock \href {https://doi.org/10.18653/v1/2020.acl-main.447} {{S}2{ORC}: The
  semantic scholar open research corpus}.
\newblock In \emph{Proceedings of the 58th Annual Meeting of the Association
  for Computational Linguistics}, pages 4969--4983, Online. Association for
  Computational Linguistics.

\bibitem[{OpenAI(2023)}]{openai2023gpt4}
OpenAI. 2023.
\newblock \href {http://arxiv.org/abs/2303.08774} {Gpt-4 technical report}.

\bibitem[{Ouyang et~al.(2022)Ouyang, Wu, Jiang, Almeida, Wainwright, Mishkin,
  Zhang, Agarwal, Slama, Gray, Schulman, Hilton, Kelton, Miller, Simens,
  Askell, Welinder, Christiano, Leike, and Lowe}]{ouyang2022training}
Long Ouyang, Jeffrey Wu, Xu~Jiang, Diogo Almeida, Carroll Wainwright, Pamela
  Mishkin, Chong Zhang, Sandhini Agarwal, Katarina Slama, Alex Gray, John
  Schulman, Jacob Hilton, Fraser Kelton, Luke Miller, Maddie Simens, Amanda
  Askell, Peter Welinder, Paul Christiano, Jan Leike, and Ryan Lowe. 2022.
\newblock \href {https://openreview.net/forum?id=TG8KACxEON} {Training language
  models to follow instructions with human feedback}.
\newblock In \emph{Advances in Neural Information Processing Systems}.

\bibitem[{Pang et~al.(2022)Pang, Parrish, Joshi, Nangia, Phang, Chen,
  Padmakumar, Ma, Thompson, He, and Bowman}]{pang-etal-2022-quality}
Richard~Yuanzhe Pang, Alicia Parrish, Nitish Joshi, Nikita Nangia, Jason Phang,
  Angelica Chen, Vishakh Padmakumar, Johnny Ma, Jana Thompson, He~He, and
  Samuel Bowman. 2022.
\newblock \href {https://doi.org/10.18653/v1/2022.naacl-main.391} {{Q}u{ALITY}:
  Question answering with long input texts, yes!}
\newblock In \emph{Proceedings of the 2022 Conference of the North American
  Chapter of the Association for Computational Linguistics: Human Language
  Technologies}, pages 5336--5358, Seattle, United States. Association for
  Computational Linguistics.

\bibitem[{Phang et~al.(2022)Phang, Zhao, and Liu}]{phang2022investigating}
Jason Phang, Yao Zhao, and Peter~J. Liu. 2022.
\newblock \href {http://arxiv.org/abs/2208.04347} {Investigating efficiently
  extending transformers for long input summarization}.

\bibitem[{Radford et~al.(2019)Radford, Wu, Child, Luan, Amodei, Sutskever
  et~al.}]{radford2019language}
Alec Radford, Jeffrey Wu, Rewon Child, David Luan, Dario Amodei, Ilya
  Sutskever, et~al. 2019.
\newblock Language models are unsupervised multitask learners.
\newblock \emph{OpenAI blog}, 1(8):9.

\bibitem[{Raffel et~al.(2020)Raffel, Shazeer, Roberts, Lee, Narang, Matena,
  Zhou, Li, and Liu}]{JMLR:v21:20-074}
Colin Raffel, Noam Shazeer, Adam Roberts, Katherine Lee, Sharan Narang, Michael
  Matena, Yanqi Zhou, Wei Li, and Peter~J. Liu. 2020.
\newblock \href {http://jmlr.org/papers/v21/20-074.html} {Exploring the limits
  of transfer learning with a unified text-to-text transformer}.
\newblock \emph{J. Mach. Learn. Res.}, 21:140:1--140:67.

\bibitem[{Rajpurkar et~al.(2016)Rajpurkar, Zhang, Lopyrev, and
  Liang}]{rajpurkar-etal-2016-squad}
Pranav Rajpurkar, Jian Zhang, Konstantin Lopyrev, and Percy Liang. 2016.
\newblock \href {https://doi.org/10.18653/v1/D16-1264} {{SQ}u{AD}: 100,000+
  questions for machine comprehension of text}.
\newblock In \emph{Proceedings of the 2016 Conference on Empirical Methods in
  Natural Language Processing}, pages 2383--2392, Austin, Texas. Association
  for Computational Linguistics.

\bibitem[{Sanh et~al.(2022)Sanh, Webson, Raffel, Bach, Sutawika, Alyafeai,
  Chaffin, Stiegler, Raja, Dey, Bari, Xu, Thakker, Sharma, Szczechla, Kim,
  Chhablani, Nayak, Datta, Chang, Jiang, Wang, Manica, Shen, Yong, Pandey,
  Bawden, Wang, Neeraj, Rozen, Sharma, Santilli, F{\'{e}}vry, Fries, Teehan,
  Scao, Biderman, Gao, Wolf, and Rush}]{sanh2022multitask}
Victor Sanh, Albert Webson, Colin Raffel, Stephen~H. Bach, Lintang Sutawika,
  Zaid Alyafeai, Antoine Chaffin, Arnaud Stiegler, Arun Raja, Manan Dey,
  M~Saiful Bari, Canwen Xu, Urmish Thakker, Shanya~Sharma Sharma, Eliza
  Szczechla, Taewoon Kim, Gunjan Chhablani, Nihal~V. Nayak, Debajyoti Datta,
  Jonathan Chang, Mike~Tian{-}Jian Jiang, Han Wang, Matteo Manica, Sheng Shen,
  Zheng~Xin Yong, Harshit Pandey, Rachel Bawden, Thomas Wang, Trishala Neeraj,
  Jos Rozen, Abheesht Sharma, Andrea Santilli, Thibault F{\'{e}}vry, Jason~Alan
  Fries, Ryan Teehan, Teven~Le Scao, Stella Biderman, Leo Gao, Thomas Wolf, and
  Alexander~M. Rush. 2022.
\newblock \href {https://openreview.net/forum?id=9Vrb9D0WI4} {Multitask
  prompted training enables zero-shot task generalization}.
\newblock In \emph{The Tenth International Conference on Learning
  Representations, {ICLR} 2022, Virtual Event, April 25-29, 2022}.
  OpenReview.net.

\bibitem[{Schick and
  Sch{\"u}tze(2021{\natexlab{a}})}]{schick-schutze-2021-exploiting}
Timo Schick and Hinrich Sch{\"u}tze. 2021{\natexlab{a}}.
\newblock \href {https://doi.org/10.18653/v1/2021.eacl-main.20} {Exploiting
  cloze-questions for few-shot text classification and natural language
  inference}.
\newblock In \emph{Proceedings of the 16th Conference of the European Chapter
  of the Association for Computational Linguistics: Main Volume}, pages
  255--269, Online. Association for Computational Linguistics.

\bibitem[{Schick and
  Sch{\"u}tze(2021{\natexlab{b}})}]{schick-schutze-2021-shot}
Timo Schick and Hinrich Sch{\"u}tze. 2021{\natexlab{b}}.
\newblock \href {https://doi.org/10.18653/v1/2021.emnlp-main.32} {Few-shot text
  generation with natural language instructions}.
\newblock In \emph{Proceedings of the 2021 Conference on Empirical Methods in
  Natural Language Processing}, pages 390--402, Online and Punta Cana,
  Dominican Republic. Association for Computational Linguistics.

\bibitem[{Shaham et~al.(2022)Shaham, Segal, Ivgi, Efrat, Yoran, Haviv, Gupta,
  Xiong, Geva, Berant, and Levy}]{shaham-etal-2022-scrolls}
Uri Shaham, Elad Segal, Maor Ivgi, Avia Efrat, Ori Yoran, Adi Haviv, Ankit
  Gupta, Wenhan Xiong, Mor Geva, Jonathan Berant, and Omer Levy. 2022.
\newblock \href {https://aclanthology.org/2022.emnlp-main.823} {{SCROLLS}:
  Standardized {C}ompa{R}ison over long language sequences}.
\newblock In \emph{Proceedings of the 2022 Conference on Empirical Methods in
  Natural Language Processing}, pages 12007--12021, Abu Dhabi, United Arab
  Emirates. Association for Computational Linguistics.

\bibitem[{Srivastava et~al.(2022)Srivastava, Rastogi, Rao, Shoeb, Abid, Fisch,
  Brown, Santoro, Gupta, and et~al.}]{srivastava2022imitation}
Aarohi Srivastava, Abhinav Rastogi, Abhishek Rao, Abu Awal~Md Shoeb, Abubakar
  Abid, Adam Fisch, Adam~R. Brown, Adam Santoro, Aditya Gupta, and Adrià
  Garriga-Alonso et~al. 2022.
\newblock \href {http://arxiv.org/abs/2206.04615} {Beyond the imitation game:
  Quantifying and extrapolating the capabilities of language models}.

\bibitem[{Tay et~al.(2023)Tay, Dehghani, Tran, Garcia, Wei, Wang, Chung, Bahri,
  Schuster, Zheng, Zhou, Houlsby, and Metzler}]{tay2023ul}
Yi~Tay, Mostafa Dehghani, Vinh~Q. Tran, Xavier Garcia, Jason Wei, Xuezhi Wang,
  Hyung~Won Chung, Dara Bahri, Tal Schuster, Steven Zheng, Denny Zhou, Neil
  Houlsby, and Donald Metzler. 2023.
\newblock \href {https://openreview.net/forum?id=6ruVLB727MC} {{UL}2: Unifying
  language learning paradigms}.
\newblock In \emph{The Eleventh International Conference on Learning
  Representations}.

\bibitem[{Trivedi et~al.(2022)Trivedi, Balasubramanian, Khot, and
  Sabharwal}]{trivedi-etal-2022-musique}
Harsh Trivedi, Niranjan Balasubramanian, Tushar Khot, and Ashish Sabharwal.
  2022.
\newblock \href {https://doi.org/10.1162/tacl_a_00475} {♫ {M}u{S}i{Q}ue:
  Multihop questions via single-hop question composition}.
\newblock \emph{Transactions of the Association for Computational Linguistics},
  10:539--554.

\bibitem[{Wang et~al.(2022)Wang, Pang, Chen, Phang, and
  Bowman}]{wang-etal-2022-squality}
Alex Wang, Richard~Yuanzhe Pang, Angelica Chen, Jason Phang, and Samuel~R.
  Bowman. 2022.
\newblock \href {https://aclanthology.org/2022.emnlp-main.75} {{SQ}u{ALITY}:
  Building a long-document summarization dataset the hard way}.
\newblock In \emph{Proceedings of the 2022 Conference on Empirical Methods in
  Natural Language Processing}, pages 1139--1156, Abu Dhabi, United Arab
  Emirates. Association for Computational Linguistics.

\bibitem[{Wei et~al.(2022{\natexlab{a}})Wei, Bosma, Zhao, Guu, Yu, Lester, Du,
  Dai, and Le}]{wei2022finetuned}
Jason Wei, Maarten Bosma, Vincent~Y. Zhao, Kelvin Guu, Adams~Wei Yu, Brian
  Lester, Nan Du, Andrew~M. Dai, and Quoc~V. Le. 2022{\natexlab{a}}.
\newblock \href {https://openreview.net/forum?id=gEZrGCozdqR} {Finetuned
  language models are zero-shot learners}.
\newblock In \emph{The Tenth International Conference on Learning
  Representations, {ICLR} 2022, Virtual Event, April 25-29, 2022}.
  OpenReview.net.

\bibitem[{Wei et~al.(2022{\natexlab{b}})Wei, Wang, Schuurmans, Bosma, ichter,
  Xia, Chi, Le, and Zhou}]{NEURIPS2022_9d560961}
Jason Wei, Xuezhi Wang, Dale Schuurmans, Maarten Bosma, brian ichter, Fei Xia,
  Ed~Chi, Quoc~V Le, and Denny Zhou. 2022{\natexlab{b}}.
\newblock \href
  {https://proceedings.neurips.cc/paper_files/paper/2022/file/9d5609613524ecf4f15af0f7b31abca4-Paper-Conference.pdf}
  {Chain-of-thought prompting elicits reasoning in large language models}.
\newblock In \emph{Advances in Neural Information Processing Systems},
  volume~35, pages 24824--24837. Curran Associates, Inc.

\bibitem[{Xiong et~al.(2022)Xiong, Gupta, Toshniwal, Mehdad, and tau
  Yih}]{xiong2022adapting}
Wenhan Xiong, Anchit Gupta, Shubham Toshniwal, Yashar Mehdad, and Wen tau Yih.
  2022.
\newblock \href {http://arxiv.org/abs/2209.10052} {Adapting pretrained
  text-to-text models for long text sequences}.

\bibitem[{Zhong et~al.(2021)Zhong, Yin, Yu, Zaidi, Mutuma, Jha, Awadallah,
  Celikyilmaz, Liu, Qiu, and Radev}]{zhong-etal-2021-qmsum}
Ming Zhong, Da~Yin, Tao Yu, Ahmad Zaidi, Mutethia Mutuma, Rahul Jha,
  Ahmed~Hassan Awadallah, Asli Celikyilmaz, Yang Liu, Xipeng Qiu, and Dragomir
  Radev. 2021.
\newblock \href {https://doi.org/10.18653/v1/2021.naacl-main.472} {{QMS}um: A
  new benchmark for query-based multi-domain meeting summarization}.
\newblock In \emph{Proceedings of the 2021 Conference of the North American
  Chapter of the Association for Computational Linguistics: Human Language
  Technologies}, pages 5905--5921, Online. Association for Computational
  Linguistics.

\end{thebibliography}

\appendix

\section{Prompts}
\label{sec:appendix}

\cref{tab:summ_prompts} shows \zs{} prompts for summarization tasks, and \cref{tab:qa_and_agg_prompts} shows our prompts for question answering and aggregation tasks. The prompts are designed to be simple, natural, and explicit. In braces are placeholders for the text of every example.

\begin{table*}[t]
    \small
    \centering
    \begin{tabular}{@{}p{0.1225\textwidth}p{0.85\textwidth}@{}}
    \toprule
    \textbf{Task} & \textbf{Prompt}\\
    \midrule
     GovReport & You are given a report by a government agency. Write a one-page summary of the report. 
     
     ~
     
     Report:
     
     \{REPORT\}
     
     ~
     
     Summary:   
     \\
     \midrule
     SummScreen & You are given a script of a TV episode. Summarize the episode in a paragraph. 
     
     ~
     
     Episode Script:
     
     \{SCRIPT\}
     
     ~
     
     Summary:   
     \\
    \midrule
     QMSum & You are given a meeting transcript and a query containing a question or instruction. Answer the query in one or more sentences.
     
     ~
     
     Transcript:
     
     \{TRANSCRIPT\}
     
     ~

     Query:

    \{QUERY\}

    ~
    
     Answer:  
     \\
    \midrule
     SQuALITY & You are given a story and a question. Answer the question in a paragraph.
     
     ~
     
     Story:
     
     \{STORY\}
     
     ~

     Question:

    \{QUESTION\}

    ~
    
     Answer:   
     \\
    \bottomrule
    \end{tabular}
    \caption{
    Summarization task prompts. For chat models (ChatGPT, Claude, and GPT-4), we and omit the response header, as it is less appropriate for dialogue.}
    \label{tab:summ_prompts}
\end{table*}

\begin{table*}[t]
    \small
    \centering
    \begin{tabular}{@{}p{0.1225\textwidth}p{0.85\textwidth}@{}}
    \toprule
    \textbf{Task} & \textbf{Prompt}\\
    \midrule
     Qasper & You are given a scientific article and a question. Answer the question as concisely as you can, using a single phrase or sentence if possible. If the question cannot be answered based on the information in the article, write "unanswerable". If the question is a yes/no question, answer "yes", "no", or "unanswerable". \textcolor[HTML]{AFAFAF}{Do not provide any explanation.}
     
     ~
     
     Article:
     
     \{ARTICLE\}
     
     ~

     Question:

    \{QUESTION\}

    ~
    
     Answer:  
     \\
    \midrule
     NarrativeQa & You are given a story, which can be either a novel or a movie script, and a question. Answer the question as concisely as you can, using a single phrase if possible. \textcolor[HTML]{AFAFAF}{Do not provide any explanation.}
     
     ~
     
     Story:
     
     \{STORY\}
     
     ~

     Question:

    \{QUESTION\}

    ~
    
     Answer:   
     \\
    \midrule
     QuALITY & You are provided a story and a multiple-choice question with 4 possible answers (marked by A, B, C, D). Choose the best answer by writing its corresponding letter (either A, B, C, or D). \textcolor[HTML]{AFAFAF}{Do not provide any explanation.}
     
     ~
     
     Story:
     
     \{STORY\}
     
     ~

     Question and Possible Answers:

    \{QUESTION\_AND\_OPTIONS\}

    ~
    
     Answer:     
     \\
    \midrule
     MuSiQue & You are given several paragraphs from Wikipedia and a question. Answer the question as concisely as you can, using a single phrase if possible. If the question cannot be answered based on the information in the paragraphs, write "unanswerable". \textcolor[HTML]{AFAFAF}{Do not provide any explanation.}
     
     ~
     
     Paragraphs:
     
     \{PARAGRAPHS\}
     
     ~

     Question:

    \{QUESTION\}

    ~
    
     Answer:   
     \\
\midrule
     SpaceDigest & You are given a list of reviews about a specific hotel. Each review is either positive or negative. What is the percentage of positive reviews (e.g. 60\%, 34\%, etc.)? \textcolor[HTML]{AFAFAF}{Do not provide any explanation.}
     
     ~
     
     Reviews:
     
     \{REVIEWS\}
     
     ~

     Percentage of Positive Reviews:  
     \\
\midrule
     BookSumSort & You are given \{NUM\_SUMMARIES\} summaries of chapters or parts of a novel, in a shuffled order, where each summary is denoted by a numerical ID (e.g. Summary 1, Summary 3, etc.). Reorder the summaries according to the original order of chapters/parts in the novel by writing a list of length \{NUM\_SUMMARIES\} of the summary IDs (e.g. if you were given 5 summaries, one possible answer could be "5, 1, 3, 4, 2"). \textcolor[HTML]{AFAFAF}{Do not provide any explanation.}
     
     ~
     
     Summaries:
     
     \{SUMMARIES\}

    ~
    
     Summary IDs in Correct Order:   
     \\
    \bottomrule
    \end{tabular}
    \caption{Question answering and aggregation task prompts. For chat models (ChatGPT, Claude, and GPT-4), we add an additional instruction (in \textcolor[HTML]{AFAFAF}{grey}), and omit the response header, as it is less appropriate for dialogue.}
    \label{tab:qa_and_agg_prompts}
\end{table*}

\end{document}